\documentclass[10pt,twocolumn,letterpaper]{article}

\usepackage{cvpr}
\usepackage{times}
\usepackage{epsfig}
\usepackage{graphicx}
\usepackage{amsmath}
\usepackage{amssymb}

\usepackage{graphicx}
\usepackage{amsmath,amssymb} 
\usepackage{makecell}
\usepackage{color}
\usepackage{enumitem}
\usepackage{booktabs}
\usepackage{float}
\usepackage{tabulary,multirow,overpic,xcolor}
\usepackage{verbatim}
\usepackage{enumitem}
\usepackage[caption=false]{subfig}
\usepackage{array}
\usepackage{arydshln}

\DeclareMathOperator*{\argmax}{arg\,max}

\newcommand{\ve}[1]{\mathbf{#1}} 
\newcommand{\ma}[1]{\mathbf{#1}} 
\newcommand{\cev}[1]{\reflectbox{\ensuremath{\vec{\reflectbox{\ensuremath{#1}}}}}}

\newcolumntype{x}[1]{>{\centering\arraybackslash}p{#1pt}}
\newcommand{\tablestyle}[2]{\setlength{\tabcolsep}{#1}\renewcommand{\arraystretch}{#2}\centering\footnotesize}

\usepackage[breaklinks=true,bookmarks=false]{hyperref}

\cvprfinalcopy 

\def\cvprPaperID{5539} 

\ifcvprfinal\fi
\begin{document}

\title{
	Local-Global Video-Text Interactions for Temporal Grounding
} 

\author{
	Jonghwan Mun$^{\rm 1,2}$ \hspace{0.5cm}
	Minsu Cho$^{\rm 1}$ \hspace{0.5cm}
	Bohyung Han$^{\rm 2}$ \\
	\hspace{-0.3cm}
	$\textsuperscript{\rm 1}$Computer Vision Lab., POSTECH, Korea \\
	$\textsuperscript{\rm 2}$Computer Vision Lab., ASRI, Seoul National University, Korea \\
	$\textsuperscript{\rm 1}${\small\texttt{\{jonghwan.mun,mscho\}@postech.ac.kr}} \hspace{0.05cm}
	$\textsuperscript{\rm 2}${\small\texttt{bhhan@snu.ac.kr}}\\
}

\maketitle
\thispagestyle{empty}

\begin{abstract}
	This paper addresses the problem of text-to-video temporal grounding, which aims to identify the time interval in a video semantically relevant to a text query. We tackle this problem using a novel regression-based model that learns to extract a collection of mid-level features for semantic phrases in a text query, which corresponds to important semantic entities described in the query (e.g., actors, objects, and actions), and reflect bi-modal interactions between the linguistic features of the query and the visual features of the video in multiple levels. The proposed method effectively predicts the target time interval by exploiting contextual information from local to global during bi-modal interactions. Through in-depth ablation studies, we find out that incorporating both local and global context in video and text interactions is crucial to the accurate grounding. Our experiment shows that the proposed method outperforms the state of the arts on Charades-STA and ActivityNet Captions datasets by large margins, 7.44\% and 4.61\% points at Recall@tIoU=0.5 metric, respectively. Code is available.\footnote{https://github.com/JonghwanMun/LGI4temporalgrounding}
\end{abstract}

\section{Introduction}
\label{sec:introduction}

As the amount of videos in the internet grows explosively, understanding and analyzing video contents (\eg, action classification~\cite{i3d,slowfast,c3d} and detection~\cite{sstat,montes2016temporal,cdc,mscnn,rc3d,yeung2016end,pgcn,ssn}) becomes increasingly important. Furthermore, with the recent advances of deep learning on top of large-scale datasets~\cite{anet, viddensecap, tvqa, marioqa}, research on video content understanding is moving towards multi-modal problems (\eg, video question answering~\cite{tvqa,marioqa}, video captioning~\cite{viddensecap,sdvc}) involving text, speech, and sound. 

\begin{figure}[!t]
	\centering
	\scalebox{1}{
		\includegraphics[width=\linewidth]{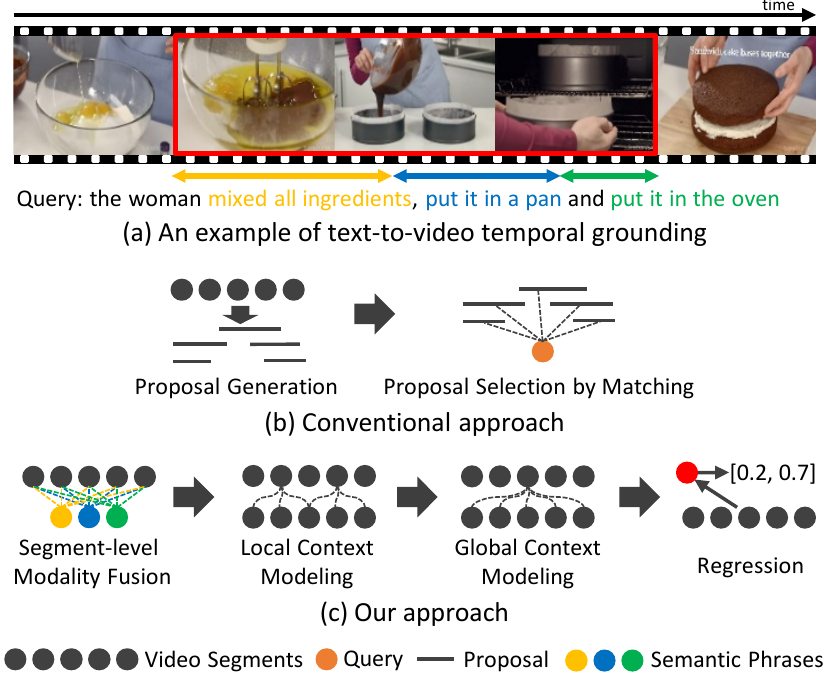}
	}
	\caption{
		Video-to-text temporal grounding. (a) An example where the target time interval (red box) consists of multiple parts related to semantic phrases in a text query.
		(b) \textit{Scan-and-localize} framework that localizes the target time interval by comparing individual proposals with the whole semantics of the query.
		(c) Our method that regresses the target time interval with the bi-modal interactions in three levels between video segments and semantic phrases identified from a query.
	}
	\label{fig:tml_eg}
\end{figure}

This paper addresses the problem of text-to-video temporal grounding, which aims to localize the time interval in a video corresponding to the expression in a text query. 
Our main idea is to extract multiple {\em semantic phrases} from the text query and align them with the video using {\em local and global interactions} between linguistic and visual features.
We define the semantic phrase as a sequence of words that may describe a semantic entity such as an actor, an object, an action, a place, etc. 
Fig.~\ref{fig:tml_eg}(a) shows an example of temporal grounding, where a text query consists of multiple semantic phrases corresponding to actors (\ie, `the woman') and actions (\ie, `mixed all ingredients', `put it in a pan', `put it in the oven'). 
This example indicates that a text query can be effectively grounded onto a video by identifying relevant semantic phrases from the query and properly aligning them with corresponding parts of the video.

Leveraging such semantic phrases of a text query, however, has never been explored in temporal grounding.
Most existing methods~\cite{mcn, tgn, sap, ctrl, acl, acrn, mlvi, man} tackle the problem typically in the \textit{scan-and-localize} framework, which in a nutshell compares a query with all candidate proposals of time intervals and selects the one with the highest matching score as shown in Fig.~\ref{fig:tml_eg}(b).
During the matching procedure, they rely on a single global feature of the query rather than finer-grained features in a phrase level, thus missing important details for localization.
Recent work~\cite{ablr} formulates the task as an attentive localization by regression and attempts to extract semantic features from a query through an attention scheme.
However, it is still limited to identifying the most discriminative semantic phrase without understanding comprehensive context.

We propose a novel regression-based method for temporal grounding as depicted in Fig.~\ref{fig:tml_eg}(c), which performs local-global video-text interactions for in-depth relationship modeling between semantic phrases and video segments.
Contrary to the existing approaches, we first extract linguistic features for semantic phrases in a query using sequential query attention.
Then, we perform video-text interaction in three levels to effectively align the semantic phrase features with segment-level visual features of a video:
1) segment-level fusion across the video segment and semantic phrase features, which highlights the segments associated with each semantic phrase,
2) local context modeling, which helps align the phrases with temporal regions of variable lengths,
and 3) global context modeling, which captures relations between phrases.
Finally, we aggregate the fused segment-level features using temporal attentive pooling and regress the time interval 
using the aggregated feature.

The main contributions are summarized as follows:
\begin{itemize}[label=$\bullet$]
	\item
	We introduce a sequential query attention module that extracts representations of multiple and distinct semantic phrases from a text query for the subsequent video-text interaction.
	\item
	We present an effective local-global video-text interaction algorithm that models the relationship between video segments and semantic phrases in multiple levels, thus enhancing final localization by regression.
	\item
	We conduct extensive experiments to validate the effectiveness of our method and show that it outperforms the state of the arts by a large margin on both Charades-STA and ActivityNet Captions datasets.
\end{itemize}

\section{Related Work}
\label{sec:related}

\subsection{Temporal Action Detection}
\label{sec:action_detection}

Recent temporal action detection methods often rely on the state-of-the-art object detection and segmentation techniques in the image domain, and can be categorized into the following three groups.
First, some methods~\cite{montes2016temporal,cdc} perform frame-level dense prediction and determine time intervals by pruning frames based on their confidence scores and grouping adjacent ones. 
Second, proposal-based techniques~\cite{mscnn,rc3d,pgcn,ssn} extract all action proposals and refine their boundaries for action detection.
Third, there exist some approaches~\cite{sstat,yeung2016end} based on single-shot detection like SSD~\cite{ssd} for fast inference.
In contrast to the action detection task, which is limited to localizing a single action instance, temporal grounding on a video by a text requires to localize more complex intervals that would involve more than two actions depending on the description in sentence queries.

\begin{figure*}[t]
	\centering
	\scalebox{0.96}{
		\includegraphics[width=\linewidth]{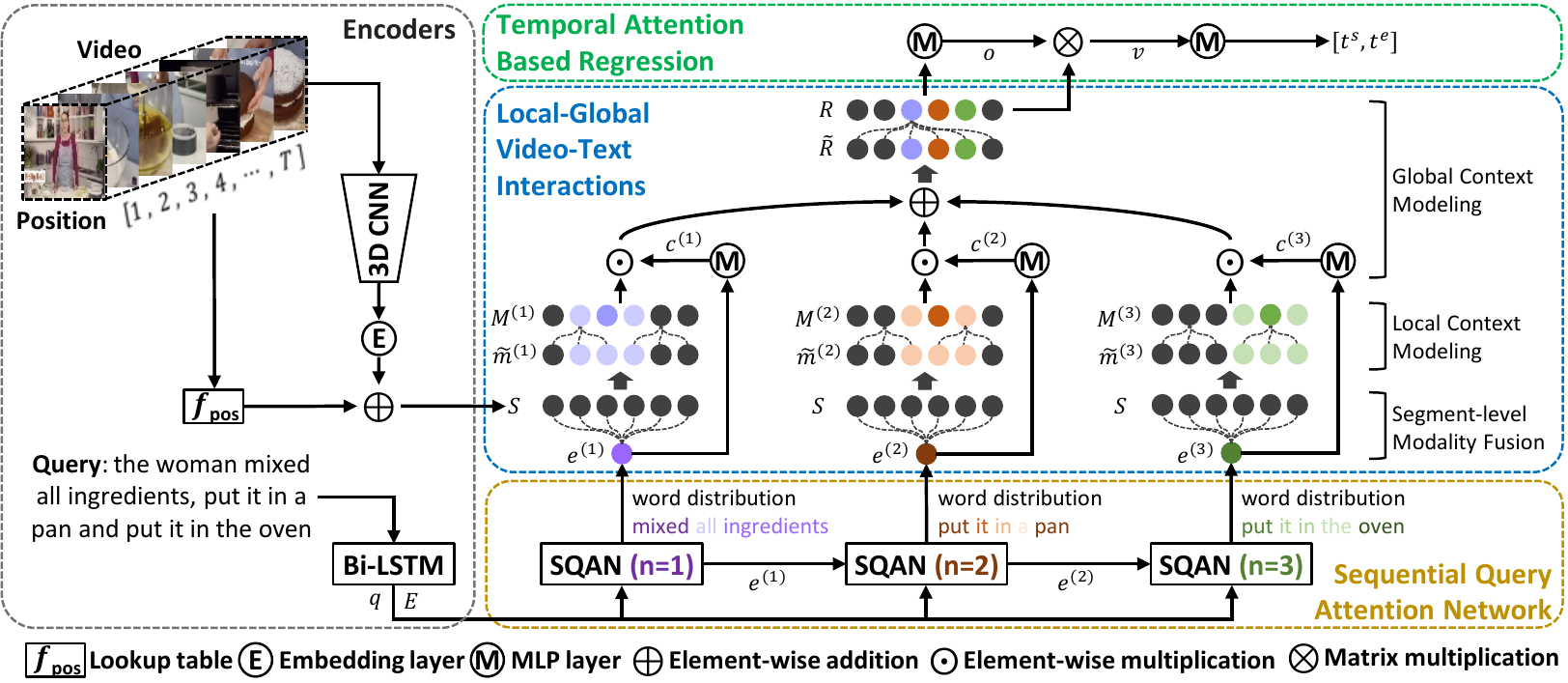}
	}
	\caption{
		Overall architecture of our algorithm.
		Given a video and a text query, we encode them to obtain segment-level visual features, word-level and sentence-level textual features (Section~\ref{sec:enc}).
		We extract a set of semantic phrase features from the query using the Sequential Query Attention Network (SQAN) (Section~\ref{sec:sqan}).
		Then, we obtain semantics-aware segment features based on the extracted phrase features via local-global video-text interactions (Section~\ref{sec:LGVTI}).
		Finally, we directly predict the time interval from the summarized video features using the temporal attention (Section~\ref{sec:locbyreg}).
		We train the model using the regression loss and two additional attention-related losses (Section~\ref{sec:training}).
	}
	\label{fig:overall_architecture}
\end{figure*}

\subsection{Text-to-Video Temporal Grounding}
\label{sec:tgld}

Since the release of two datasets for text-to-video temporal grounding, referred to as DiDeMo and Charades-STA, various algorithms~\cite{mcn,ctrl,acl,acrn,man} have been proposed within the \textit{scan-and-localize} framework, where candidate clips are obtained by scanning a whole video based on sliding windows and the best matching clip with an input text query is eventually selected.
As the sliding window scheme is time-consuming and often contains redundant candidate clips, more effective and efficient methods~\cite{tgn,sap,mlvi} are proposed as alternatives;
a LSTM-based single-stream network~\cite{tgn} is proposed to perform frame-by-word interactions and the clip proposal generation based methods~\cite{sap,mlvi} are proposed to reduce the number of redundant candidate clips.
Although those methods successfully enhance processing time, they still need to observe full videos, thus, reinforcement learning is introduced to observe only a fraction of frames~\cite{smrl} or a few clips~\cite{rwm} for temporal grounding.

On the other hand, proposal-free algorithms~\cite{excl,pftml,ablr} have also been proposed.
Inspired by the recent advance in text-based machine comprehension, Ghosh \etal~\cite{excl} propose to directly identify indices of video segments corresponding to start and end positions, and Opazo \etal~\cite{pftml} improve the method by adopting a query-guided dynamic filter.
Yuan \etal~\cite{ablr} present a co-attention based location regression algorithm, where the attention is learned to focus on video segments within ground-truth time intervals.

ABLR~\cite{ablr} is the most similar to our algorithm in the sense that it formulates the task as the attention-based location regression.
However, our approach is different from ABLR in the following two aspects.
First, ABLR focuses only on the most discriminative semantic phrase in a query to acquire visual information, whereas we consider multiple ones for more comprehensive estimation.
Second, ABLR relies on coarse interactions between video and text inputs and often fails to capture fine-grained correlations between video segments and query words. In contrast, we perform a more effective multi-level video-text interaction to model correlations between semantic phrases and video segments.

\section{Proposed Method}
\label{sec:framework}
This section describes our main idea and its implementation using a deep neural network in detail.

\subsection{Algorithm Overview}
\label{sec:overview}

Given an untrimmed video $V$, a text query $Q$ and a time interval of target region $C$ within $V$, existing methods typically learn the models parametrized by $\theta$ to maximize the following expected log-likelihood:
\begin{equation}
\theta^{*} = \argmax_{\theta} ~\mathbb{E} [\log p_{\theta} (C|V,Q)].
\label{eq:existing_objective}
\end{equation}
Note that, in the above objective function, the text query $Q$ often involves multiple semantic phrases as presented in Fig.~\ref{fig:tml_eg}(a), which requires modeling finer-level relations between a query and a video besides global ones to achieve precise localization in temporal grounding.
To realize this idea, we introduce a differentiable module $f_{\text{e}}$ to represent a query as a set of semantic phrases and incorporate local-global video-text interactions for in-depth understanding of the phrases within a video, which leads to a new objective as follows:
\begin{equation}
\theta^{*} = \argmax_{\theta} ~\mathbb{E} [\log p_{\theta} (C|V,f_{\text{e}}(Q))].
\label{eq:our_objective}
\end{equation}

Fig.~\ref{fig:overall_architecture} illustrates the overall architecture of the proposed method.
We first compute segment-level visual features combined with their embedded timestamps, and then derive word- and sentence-level features based on the query.
Next, the Sequential Query Attention Network (SQAN) extracts multiple semantic phrase features from the query by attending over word-level features sequentially.
Then, we obtain semantics-aware segment features via multi-level video-text interactions; 
the segment feature corresponding to each semantic phrase is highlighted through a segment-level modality fusion followed by local context modeling while the relations between phrases are estimated by global context modeling.
Finally, the time intervals are predicted using the temporally attended semantics-aware segment features.

\subsection{Encoders}
\label{sec:enc}

\paragraph{Query encoding}
For a text query with $L$ words, we employ a two-layer bi-directional LSTM to obtain word- and sentence-level representations, where the bi-directional LSTM is applied to the embedded word features.
A word-level feature at the $l$-th position is obtained by the concatenation of hidden states in both directions, which is given by $\ve{w}_l=[\vec{\ve{h}}_l;\cev{\ve{h}}_l]\in \mathbb{R}^{d}$, while a sentence-level feature $\ve{q}$ is provided by the concatenation of the last hidden states in both the forward and backward LSTMs, \ie, $\ve{q}=[\vec{\ve{h}}_{L}, \cev{\ve{h}}_1] \in \mathbb{R}^{d}$ where $d$ denotes feature dimension.

\vspace{-0.2cm}
\paragraph{Video encoding}
An untrimmed video is divided into a sequence of segments with a fixed length (\eg, 16 frames), where two adjacent segments overlap each other for a half of their lengths.
We extract the features from individual segments using a 3D CNN module, denoted by $f_{\text{v}}(\cdot)$, after the uniform sampling of $T$ segments, and feed the features to an embedding layer followed by a ReLU function to match their dimensions with query features.
Formally, let $\mathbf{S}=\left[ \ve{s}_1,...,\ve{s}_T\right] \in \mathbb{R}^{d\times T}$ be a matrix that stores the $T$ sampled segment features in its columns\footnote{Although semantic phrases are sometimes associated with spatio-temporal regions in a video, for computational efficiency, we only consider temporal relationship between phrases and a  video, and use spatially pooled representation for each segment.}.
If the input videos are short and the number of segments is less than $T$, the missing parts are filled with zero vectors.
We append the temporal position embedding of each segment to the corresponding segment feature vector as done in \cite{bert} to improve accuracy in practice.
This procedure leads to the following equation for video representation: 
\begin{equation}
\ma{S} = \text{ReLU}( \mathbf{W}_{\text{seg}} f_{\text{v}}(V)) + f_\text{pos}(\mathbf{W}_{\text{pos}},[1,...,T]),
\end{equation}
where $\mathbf{W}_{\text{seg}} \in \mathbb{R}^{d\times d_v}$ denotes a learnable segment feature embedding matrix while $f_\text{pos}(\cdot, \cdot)$ is a lookup table defined by an embedding matrix $\mathbf{W}_{\text{pos}} \in \mathbb{R}^{d\times T}$ and a timestamp vector $[1, \dots, T]$.
Note that $d_v$ is the dimension of feature provided by $f_{\text{v}}(\cdot)$.
Since we formulate the given task as a location regression problem, the position encoding is a crucial step for identifying semantics at diverse temporal locations in the subsequent procedure.

\subsection{Sequential Query Attention Network (SQAN)}
\label{sec:sqan}

SQAN, denoted by $f_\text{e}(\cdot)$ in Eq.~\eqref{eq:our_objective}, plays a key role in identifying semantic phrases describing semantic entities (\eg, actors,  objects, and actions) that should be observed  in videos for precise localization.
Since there is no ground-truth for semantic phrases, we learn their representations in an end-to-end manner.
To this end, we adopt an attention mechanism with an assumption that semantic phrases are defined by a sequence  of words in a query as shown in Fig.~\ref{fig:tml_eg}(a).
Those semantic phrases can be extracted independently of each other.
Note, however, that since our goal is to obtain {\em distinct} phrases, we extract them by sequentially conditioning on preceding ones as in \cite{mac,dga}.

Given $L$ word-level features $\ma{E}=[\ve{w}_1,...,\ve{w}_L] \in \mathbb{R}^{d\times L}$ and a sentence-level feature $\ve{q} \in \mathbb{R}^{d}$, we extract $N$ semantic phrase features $\{ \ve{e}^{(1)}, \dots, \ve{e}^{(N)} \}$.
In each step $n$, a guidance vector $\ve{g}^{(n)} \in \mathbb{R}^{d}$ is obtained by embedding the vector that concatenates a linearly transformed sentence-level feature and the previous semantic phrase feature $\ve{e}^{(n-1)} \in \mathbb{R}^{d}$, which is given by
\begin{equation}
\ve{g}^{(n)} = \text{ReLU}( \ma{W}_{\text{g}} ([\ma{W}_{\text{q}}^{(n)}\ve{q} ; \ve{e}^{(n-1)}])),
\end{equation}
where $\ma{W}_{\text{g}} \in \mathbb{R}^{d\times 2d}$ and $\ma{W}_{\text{q}}^{(n)} \in \mathbb{R}^{d\times d}$ are learnable embedding matrices.
Note that we use different embedding matrix $\ma{W}_{\text{q}}^{(n)}$ at each step to attend more readily to different aspects of the query.
Then, we obtain the current semantic phrase feature $\ve{e}^{(n)}$ by estimating the attention weight vector ${\ve{a}}^{(n)} \in \mathbb{R}^{L}$ over word-level features and computing a weighted sum of the word-level features as follows:
\begin{align}
{\alpha}_l^{(n)} &= \ma{W}_{\text{qatt}} (\text{tanh}(\ma{W}_{\text{g}\alpha}\ve{g}^{(n)} + \ma{W}_{\text{w}\alpha}\ve{w}_l)), \\
\ve{a}^{(n)} &= \text{softmax}([ {\alpha}_1^{(n)}, ..., {\alpha}_L^{(n)} ]), \\
\ve{e}^{(n)} &= \sum_{l=1}^L {\ve{a}}^{(n)}_l \ve{w}_l,
\end{align}
where $\ma{W}_{\text{qatt}} \in \mathbb{R}^{1\times \frac{d}{2}}$, $\ma{W}_{\text{g}\alpha} \in \mathbb{R}^{\frac{d}{2}\times d}$ and $\ma{W}_{\text{w}\alpha} \in \mathbb{R}^{\frac{d}{2}\times d}$ are learnable embedding matrices in the query attention layer, and $\alpha^{(n)}_l$ is the confidence value for the $l$-th word at the $n$-th step.

\subsection{Local-Global Video-Text Interactions}
\label{sec:LGVTI}

Given the semantic phrase features, we perform video-text interactions in three levels with two objectives: 1) individual semantic phrase understanding, and 2) relation modeling between semantic phrases.

\vspace{-0.2cm}
\paragraph{Individual semantic phrase understanding}
Each semantic phrase feature interacts with individual segment features in two levels: {\em segment-level modality fusion} and {\em local context modeling}.
During the segment-level modality fusion, we encourage the segment features relevant to the semantic phrase features to be highlighted and the irrelevant ones to be suppressed.
However, segment-level interaction is not sufficient to understand long-range semantic entities properly since each segment has a limited field-of-view of 16 frames.
We thus introduce the local context modeling that considers neighborhood of individual segments.

With this in consideration, we perform the segment-level modality fusion similar to \cite{hadamard} using the Hadamard product while modeling the local context based on a residual block (ResBlock) that consists of two temporal convolution layers.
Note that we use kernels of large bandwidth (\eg, 15) in the ResBlock to cover long-range semantic entities.
The whole process is summarized as follows:
\begin{align}
\ve{\tilde{\ve{m}}}^{(n)}_i &= \ma{W}^{(n)}_{\text{m}}( \ma{W}^{(n)}_{\text{s}}\ve{s}_i \odot \ma{W}^{(n)}_{\text{e}}\ve{e}^{(n)} ), \\
\ma{M}^{(n)} &= \text{ResBlock} ([\tilde{\ve{m}}^{(n)}_1,...,\tilde{\ve{m}}^{(n)}_T]),
\end{align}
where $\ma{W}^{(n)}_{\text{m}} \in \mathbb{R}^{d\times d}$, $\ma{W}^{(n)}_{\text{s}} \in \mathbb{R}^{d\times d}$ and $\ma{W}^{(n)}_{\text{e}} \in \mathbb{R}^{d\times d}$ are learnable embedding matrices for segment-level fusion, and $\odot$ is the Hadamard product operator.
Note that $\tilde{\ve{m}}^{(n)}_i \in \mathbb{R}^{d}$ stands for the $i$-th bi-modal segment feature after segment-level fusion, and $\ma{M}^{(n)} \in \mathbb{R}^{d\times T}$ denotes a semantics-specific segment feature for the $n$-th semantic phrase feature $\ve{e}^{(n)}$.

\vspace{-0.2cm}
\paragraph{Relation modeling between semantic phrases}
After obtaining a set of $N$ semantics-specific segment features, $\{\ma{M}^{(1)},...,\ma{M}^{(N)}\}$, independently,  we take contextual and temporal relations between semantic phrases into account.
For example, in Fig.~\ref{fig:tml_eg}(a), understanding `it' in a semantic phrase of `put it in a pan' requires the context from another phrase of `mixed all ingredients.'
Since such relations can be defined between semantic phrases with a large temporal gap, we perform {\em global context modeling} by observing all the other segments.

For the purpose, we first aggregate $N$ segment features specific to semantic phrases, $\{\ma{M}^{(1)},...,\ma{M}^{(N)}\}$, using attentive pooling, where the weights are computed based on the corresponding semantic phrase features, as shown in Eq.~\eqref{eq:sattw} and \eqref{eq:satt}.
Then, we employ Non-Local block~\cite{nlblock} (NLBlock) that is widely used to capture global context.
The process of global context modeling is given by
\begin{align}
\ve{c} &= \text{softmax}  (\text{MLP}_{\text{satt}}( [\ve{e}^{(1)}, ..., \ve{e}^{(N)}] )), \label{eq:sattw}\\
\tilde{\ma{R}} &= \sum_{n=1}^{N} \ve{c}^{(n)} \ma{M}^{(n)}, \label{eq:satt}\\
\ma{R} &= \text{NLBlock}(\tilde{\ma{R}}) \\ 
&= \tilde{\ma{R}} + (\ma{W}_{\text{rv}}\tilde{\ma{R})}~
{\text{softmax} \left( \frac{ (\ma{W}_{\text{rq}}\tilde{\ma{R}})^{\mathsf{T}} (\ma{W}_{\text{rk}}\tilde{\ma{R}})  }{\sqrt{d}}\right)}^{\mathsf{T}}
\nonumber,
\end{align}
where $\text{MLP}_{\text{satt}}$ denotes a multilayer perceptron (MLP) with a hidden layer of $\frac{d}{2}$-dimension and $\ve{c} \in \mathbb{R}^{N}$ is a weight vector for the $N$ semantics-specific segment features.
$\tilde{\ma{R}} \in \mathbb{R}^{d\times T}$ is the aggregated feature via attentive pooling, and $\ma{R} \in \mathbb{R}^{d\times T}$ is the final semantics-aware segment features using the proposed local-global video-text interactions.
Note that $\ma{W}_{\text{rq}} \in \mathbb{R}^{d\times d}$, $\ma{W}_{\text{rk}}  \in \mathbb{R}^{d\times d}$ and $\ma{W}_{\text{rv}} \in \mathbb{R}^{d\times d}$ are learnable embedding matrices in the NLBlock.

\subsection{Temporal Attention based Regression}
\label{sec:locbyreg}
Once the semantics-aware segment features are obtained, we summarize the information while highlighting important segment features using temporal attention, and finally predict the time interval ($t^s$, $t^e$) using an MLP as follows:
\begin{align}
\ve{o} &= \text{softmax} (\text{MLP}_{\text{tatt}}(\ma{R})), \\
\ve{v} &= \sum_{i=1}^T \ve{o}_i \ma{R}_i, \\
t^s, t^e &= \text{MLP}_{\text{reg}}(\ve{v}),
\label{eq:regression}
\end{align}
where $\ve{o} \in \mathbb{R}^{T}$ and $\ve{v} \in \mathbb{R}^{d}$ are attention weights for segments and summarized video feature, respectively.
Note that $\text{MLP}_{\text{tatt}}$ and $\text{MLP}_{\text{reg}}$ have $\frac{d}{2}$- and $d$-dimensional hidden layers, respectively.

\subsection{Training}
\label{sec:training}

We train the network using three loss terms---1) location regression loss $\mathcal{L}_{\text{reg}}$, 2) temporal attention guidance loss $\mathcal{L}_{\text{tag}}$, and 3) distinct query attention loss $\mathcal{L}_{\text{dqa}}$, and the total loss is given by
\begin{equation}
\mathcal{L} = \mathcal{L}_{\text{reg}} + \mathcal{L}_{\text{tag}} + \mathcal{L}_{\text{dqa}}.
\label{eq:all_loss}
\end{equation}

\paragraph{Location regression loss}
Following \cite{ablr}, the regression loss is defined as the sum of smooth $\text{L}_1$ distances between the normalized ground-truth time interval $(\hat{t}^s, \hat{t}^e) \in [0,1]$ and our prediction $(t^s, t^e)$ as follows:
\begin{equation}
\mathcal{L}_{\text{reg}}= \text{smooth}_{\text{L}1}(\hat{t}^s - t^s ) + \text{smooth}_{\text{L}1}(\hat{t}^e - t^e),
\label{eq:reg_loss}
\end{equation}
where $\text{smooth}_{\text{L}1}(x)$ is defined as $0.5x^2$ if $|x|<1$ and $|x|-0.5$ otherwise.

\begin{figure}[!t]
	\centering
	\scalebox{0.98}{
		\includegraphics[width=0.5\linewidth]{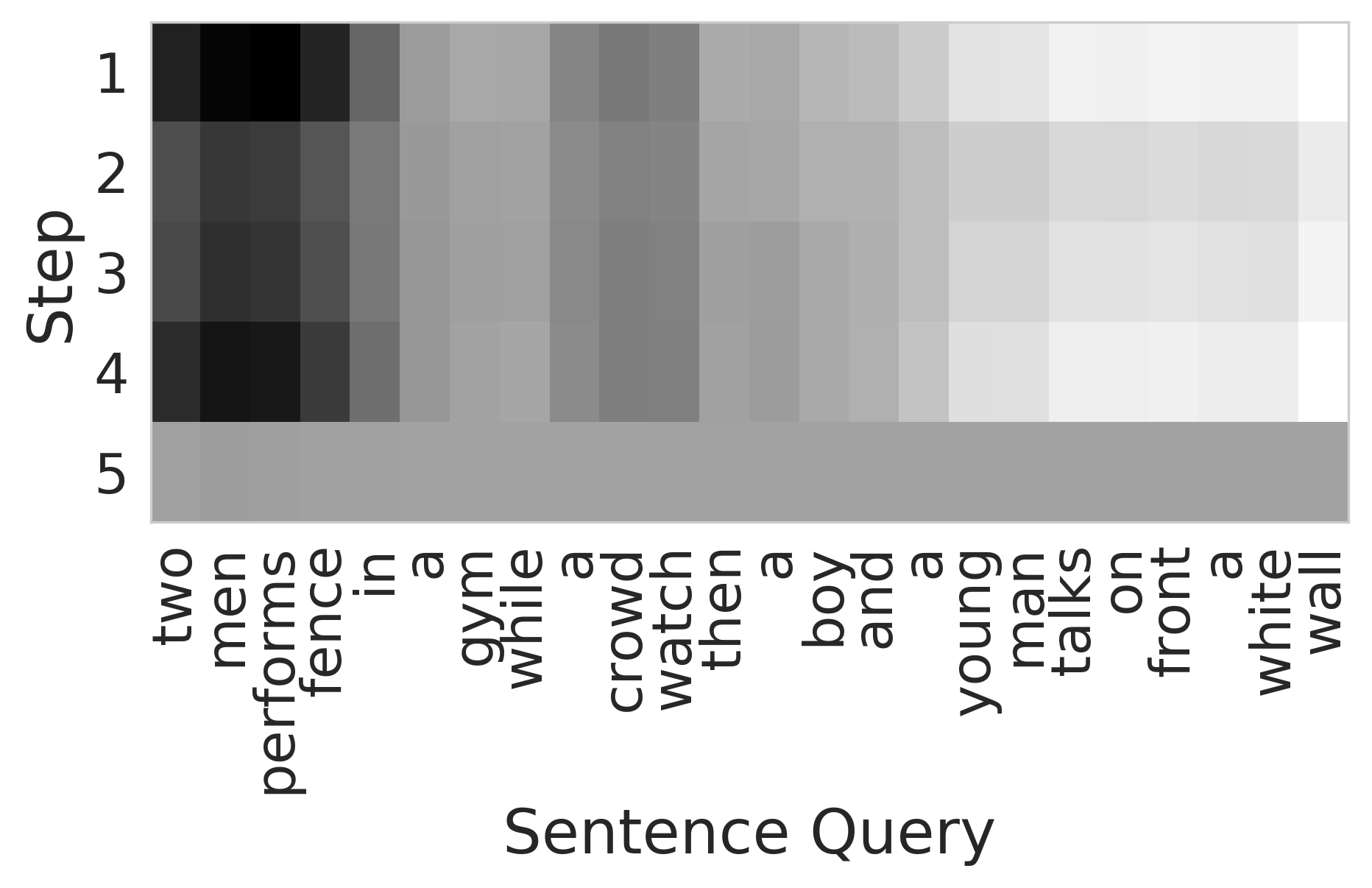}
		\includegraphics[width=0.5\linewidth]{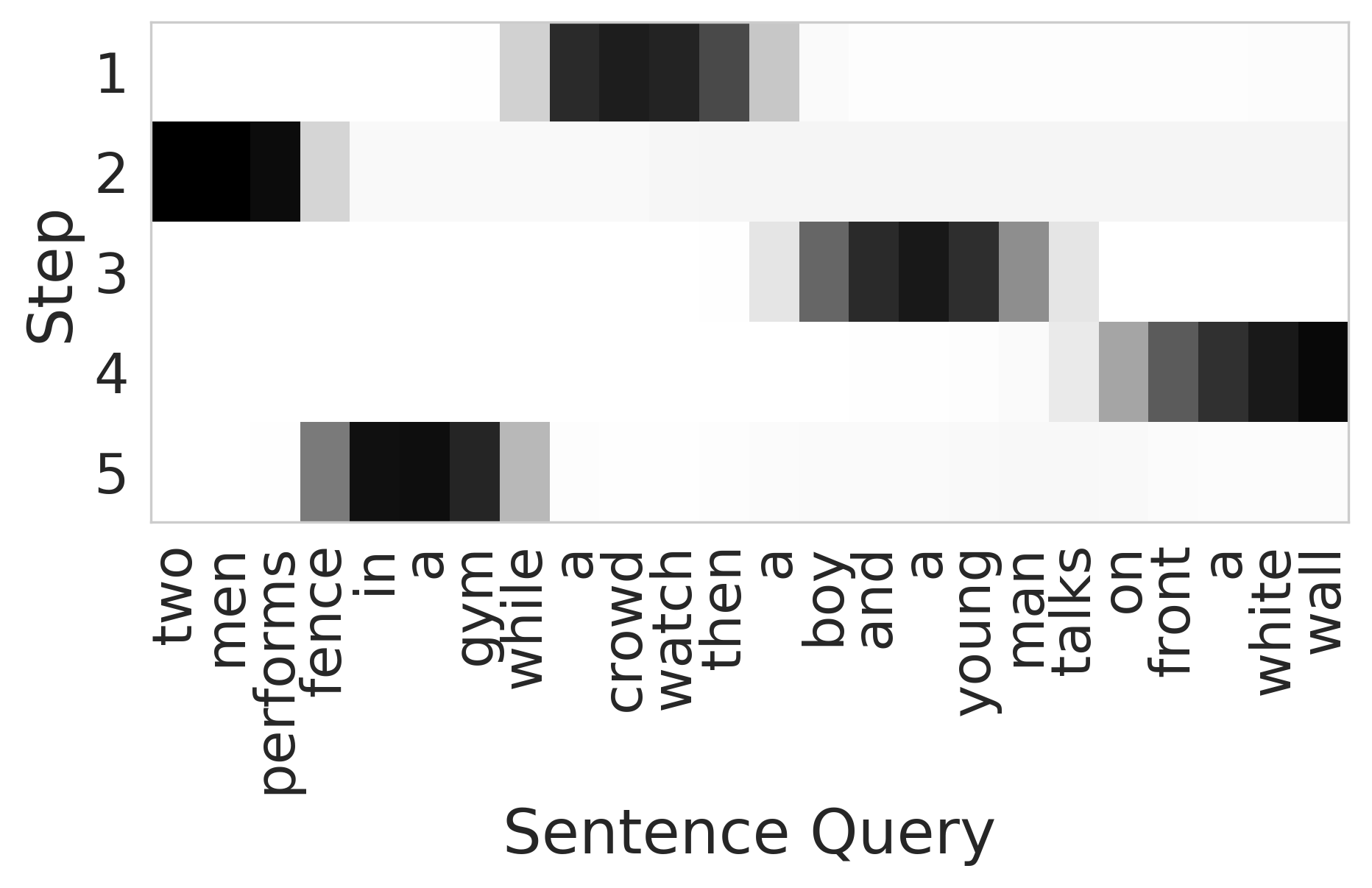}
	}
	\caption{
		Visualization of query attention weights (left) without the distinct query attention loss and (right) with it.
		SQAN successfully extracts semantic phrases corresponding to actors and actions across different steps.
	}
	\label{fig:comp_qatt}
\end{figure}

\paragraph{Temporal attention guidance loss}
Since we directly regress the temporal positions from temporally attentive features, the quality of temporal attention is critical.
Therefore, we adopt the temporal attention guidance loss proposed in \cite{ablr}, which is given by
\begin{equation}
\mathcal{L}_{\text{tag}} = -\frac{\sum_{i=1}^T \hat{\ve{o}}_i \log(\ve{o}_i)}{\sum_{i=1}^T \hat{\ve{o}}_i},
\label{eq:tag_loss}
\end{equation}
where $\hat{\ve{o}_i}$ is set to 1 if the $i$-th segment is located within the ground-truth time interval and 0 otherwise.
The attention guidance loss makes the model obtain higher attention weights for the segments related to the text query.

\paragraph{Distinct query attention loss}
Although SQAN is designed to capture different semantic phrases in a query, we observe that the query attention weights in different steps are often similar as depicted in Fig.~\ref{fig:comp_qatt}.
Thus, 
we adopt a regularization term introduced in \cite{lin2017structured} to enforce query attention weights to be distinct along different steps:
\begin{equation}
\mathcal{L}_{\text{dqa}}= ||(\ma{A}^{\mathsf{T}}\ma{A})-\lambda I||^2_F,
\label{eq:dqa_loss}
\end{equation}
where $\ma{A} \in \mathbb{R}^{L\times N}$ is the concatenated query attention weights across $N$ steps and $||\cdot||_F$ denotes Frobenius norm of a matrix.
The loss encourages attention distributions to have less overlap by making the query attention weights at two different steps decorrelated.
Note that $\lambda \in [0,1]$ controls the extent of overlap between query attention distributions;
when $\lambda$ is close to 1, the attention weights are learned to be the one-hot vector.
Fig.~\ref{fig:comp_qatt} clearly shows that the regularization term encourages the model to focus on distinct semantic phrases across query attention steps.

\section{Experiments}
\label{sec:experiments}
\subsection{Datasets}
\label{sec:datasets}

\paragraph{Charades-STA}
The dataset is collected from the Charades dataset for evaluating text-to-video temporal grounding by \cite{ctrl}, which is composed of 12,408 and 3,720 time interval and text query pairs in training and test set, respectively.
The videos are 30 seconds long on average and the maximum length of a text query is set to 10.
\vspace{-0.5cm}
\paragraph{ActivityNet Captions} 
This dataset, which has originally been constructed for dense video captioning, consists of 20k YouTube videos with an average length of 120 seconds.
It is divided into 10,024, 4,926, and 5,044 videos for training, validation, and testing, respectively.
The videos contain 3.65 temporally localized time intervals and sentence descriptions on average, where the average length of the descriptions is 13.48 words.
Following the previous methods, we report the performance of our algorithm on the combined two validation set (denoted by \textit{val\_1} and \textit{val\_2}) since annotations of the test split is not publicly available.

\begin{table}[t]
	\centering
	\caption{
		Performance comparison with other algorithms on the Charades-STA dataset.
		The bold-faced numbers mean the best performance.
	}
	\vspace{0.1cm}
	\scalebox{0.9}{
		\begin{tabular}{c|cccc}
			\toprule
			Method & R@0.3 & R@0.5 & R@0.7 & mIoU \\
			\hline\hline 
			Random   &    -      &   8.51  &   3.03  &  -  \\
			CTRL~\cite{ctrl}     &    -    &  21.42  &   7.15  &  -  \\
			SMRL~\cite{smrl}     &    -    &  24.36  &   9.01  &  -  \\
			SAP~\cite{sap}       &    -    &  27.42  &  13.36  &  -  \\
			ACL~\cite{acl}       &    -    &  30.48  &  12.20  &  -  \\
			MLVI~\cite{mlvi}     &  54.70  &  35.60  &  15.80  &  -  \\
			TripNet~\cite{tripnet} &  51.33  &  36.61  &  14.50  &  -  \\
			RWM~\cite{rwm}         &    -    &  36.70  &    -    &  -  \\
			ExCL~\cite{excl}       &  65.10  &  44.10  &  22.60  &  -  \\
			MAN~\cite{man}         &  -      &  46.53  &  22.72  &  -  \\
			PfTML-GA~\cite{pftml}  &  67.53      &  52.02  &  33.74  &  -  \\
			\hline
			
			Ours & \textbf{72.96}  &  \textbf{59.46}  &  \textbf{35.48}  &  \textbf{51.38}  \\
			\bottomrule
		\end{tabular}
	}
	\label{tab:exp_charades}
	\vspace{-0.2cm}
\end{table}

\subsection{Metrics}
\label{sec:metrics}

Following \cite{ctrl}, we adopt two metrics for the performance comparison:
1) Recall at various thresholds of the temporal Intersection over Union (R@tIoU) to measure the percentage of predictions that have tIoU with ground-truth larger than the thresholds, and 2) mean averaged tIoU (mIoU).
We use three tIoU threshold values, $\{0.3, 0.5, 0.7\}$.

\subsection{Implementation Details}
\label{sec:implementation_details}

For the 3D CNN modules to extract segment features for Charades-STA and ActivityNet Captions datasets, we employ I3D~\cite{i3d}~\footnote{https://github.com/piergiaj/pytorch-i3d} and C3D~\cite{c3d}~\footnote{http://activity-net.org/challenges/2016/download.html\#c3d} networks, respectively, while fixing their parameters during a training step.
We uniformly sample $T~(=128)$ segments from each video. 
For query encoding, we maintain all word tokens after lower-case conversion and tokenization;
vocabulary sizes are 1,140 and 11,125 for Charades-STA and ActivityNet Captions datasets, respectively.
We truncate all text queries that have maximum 25 words for ActivityNet Captions dataset.
For sequential query attention network, we extract 3 and 5 semantic phrases and set $\lambda$ in Eq.~\eqref{eq:dqa_loss} to 0.3 and 0.2 for Charades and ActivityNet Captions datasets, respectively.
In all experiments, we use Adam~\cite{kingma2014adam} to learn models with a mini-batch of 100 video-query pairs and a fixed learning rate of 0.0004.
The feature dimension $d$ is set to 512.

\begin{table}[t]
	\centering
	\caption{
		Performance comparison with other algorithms on the ActivityNet Captions dataset.
		The bold-faced numbers denote the best performance.
	}
	\vspace{0.1cm}
	\scalebox{0.9}{
		\begin{tabular}{c|cccc}
			\toprule
			Method & R@0.3 & R@0.5 & R@0.7 & mIoU \\
			\hline\hline
			MCN~\cite{mcn}       &  21.37  &   9.58  &    -    &  15.83  \\
			CTRL~\cite{ctrl}     &  28.70  &  14.00  &    -    &  20.54  \\
			ACRN~\cite{acrn}     &  31.29  &  16.17  &    -    &  24.16  \\
			MLVI~\cite{mlvi}     &  45.30  &  27.70  &  13.60  &    -    \\
			TGN~\cite{tgn}       &  45.51  &  28.47  &    -    &    -    \\
			TripNet~\cite{tripnet} &  45.42  &  32.19  &  13.93  &    -    \\
			PfTML-GA~\cite{pftml}  &  51.28  &  33.04  &  19.26  &  37.78  \\
			ABLR~\cite{ablr}       &  55.67  &  36.79  &    -    &  36.99  \\
			RWM~\cite{rwm}         &  -      &  36.90  &    -    &    -    \\
			
			\hline
			Ours    &  \textbf{58.52}  &  \textbf{41.51}  &  \textbf{23.07}  &  \textbf{41.13}  \\
			\bottomrule
		\end{tabular}
	}
	\label{tab:exp_anet}
	\vspace{-0.2cm}
\end{table}

\begin{table*}[!t]
	\centering
	\caption{
		Results of main ablation studies on the Charades-STA dataset.
		The bold-faced numbers means the best performance.
	}
	\vspace{0.1cm}
	\scalebox{0.9}{
		\begin{tabular}{l|cccc|cccc}
			\toprule
			\multirow{2}{*}{Method} & \multicolumn{2}{c}{Query Information}    & \multicolumn{2}{|c|}{Loss Terms}  & \multirow{2}{*}{R@0.3}& \multirow{2}{*}{R@0.5} & \multirow{2}{*}{R@0.7} & \multirow{2}{*}{mIoU} \\
			&  sentence ($\ve{q}$) & phrase ($\ve{e}$) & \multicolumn{1}{|c}{$+\mathcal{L}_{\text{tag}}$} & $+\mathcal{L}_{\text{dqa}}$ & & & & \\
			\hline\hline 
			
			LGI   & & \multicolumn{1}{c|}{$\surd$} & $\surd$ & $\surd$ & \textbf{72.96}  &  \textbf{59.46}  &  \textbf{35.48}  &  \textbf{51.38}  \\
			LGI w/o $\mathcal{L}_{\text{dqa}}$       & & \multicolumn{1}{c|}{$\surd$} & $\surd$ & & 71.42  &  58.28  &  34.30  &  50.24  \\
			LGI w/o $\mathcal{L}_{\text{tag}}$ & & \multicolumn{1}{c|}{$\surd$} & & $\surd$ & 61.91  &  47.12  &  24.62  &  42.43  \\
			\hdashline
			LGI--SQAN & $\surd$ & \multicolumn{1}{c|}{} & $\surd$ & &  71.02  &  57.34  &  33.25  &  49.52  \\
			LGI--SQAN  w/o $\mathcal{L}_{\text{tag}}$ & $\surd$ & \multicolumn{1}{c|}{} & & &  57.66  &  43.33  &  22.74  &  39.53  \\	
			\bottomrule
		\end{tabular}
	}
	\label{tab:main_ablation}
	\vspace{-0.2cm}
\end{table*}

\subsection{Comparison with Other Methods}
\label{sec:comparison_others}

We compare our algorithm with several recent methods, which are divided into two groups: \textit{scan-and-localize} methods, which include MCN~\cite{mcn}, CTRL~\cite{ctrl}, SAP~\cite{sap}, ACL~\cite{acl}, ACRN~\cite{acrn}, MLVI~\cite{mlvi}, TGN~\cite{tgn}, MAN~\cite{man}, TripNet~\cite{tripnet}, SMRL~\cite{smrl}, and RWM~\cite{rwm}, and proposal-free algorithms such as ABLR~\cite{ablr}, ExCL~\cite{excl}, and PfTML-GA~\cite{pftml}.

Table~\ref{tab:exp_charades} and Table~\ref{tab:exp_anet} summarize the results on Charades-STA and ActivityNet Captions datasets, respectively, where our algorithm outperforms all competing methods.
It is noticeable that the proposed technique surpasses the state-of-the-art performances by 7.44\% and 4.61\% points in terms of R@0.5 metric, respectively.

\subsection{In-Depth Analysis}
\label{sec:ablation}

For a better understanding of our algorithm, we analyze the contribution of the individual components.

\subsubsection{Main Ablation Studies}
We first investigate the contribution of sequential query attention network (SQAN) and loss terms on the Charades-STA dataset.
In this experiment, we train five variants of our model: 
1) LGI: our full model performing local-global video-text interactions based on the extracted semantic phrase features by SQAN and being learned using all loss terms,
2) LGI w/o $\mathcal{L}_{\text{dqa}}$: LGI learned without distinct query attention loss $\mathcal{L}_{\text{dqa}}$, 
3) LGI w/o $\mathcal{L}_{\text{tag}}$: LGI learned without temporal attention guidance loss $\mathcal{L}_{\text{tag}}$, 
4) LGI--SQAN: a model localizing a text query with sentence-level feature $\ve{q}$ without SQAN,
5) LGI--SQAN w/o $\mathcal{L}_{\text{tag}}$: LGI--SQAN learned without $\mathcal{L}_{\text{tag}}$.
Note that the architecture of LGI--SQAN is depicted in supplementary material.

\begin{figure}[!t]
	\centering
	\begin{minipage}[t]{0.48\linewidth}
		\centering
		\subfloat[Charades-STA \label{tab:ablation:natt_charades}]{
			\centering
			\includegraphics[width=\linewidth]{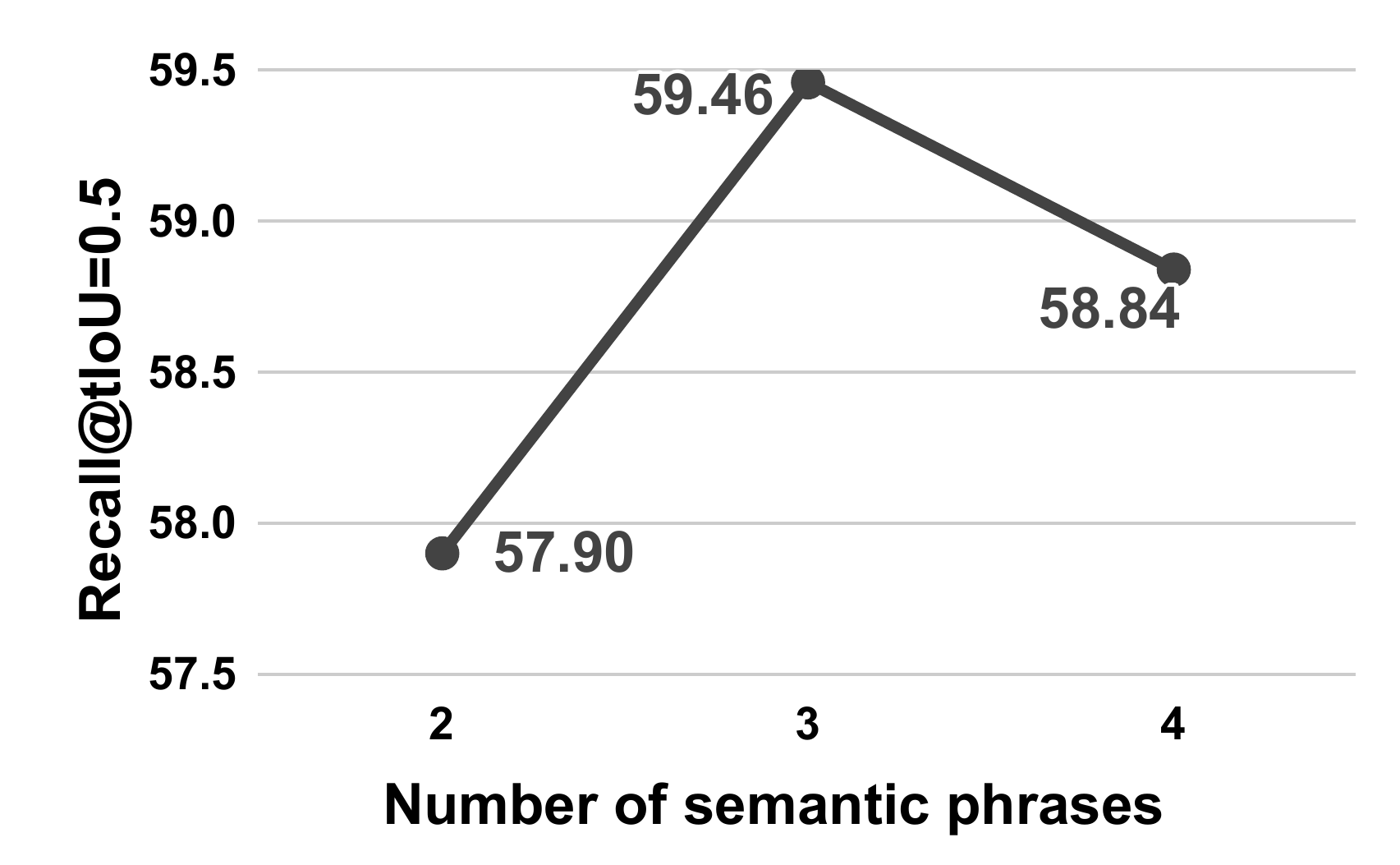}
		}
	\end{minipage}
	\begin{minipage}[t]{0.48\linewidth}
		\centering		
		\subfloat[ActivityNet Captions \label{tab:ablation:natt_anet}]{
			\centering
			\includegraphics[width=\linewidth]{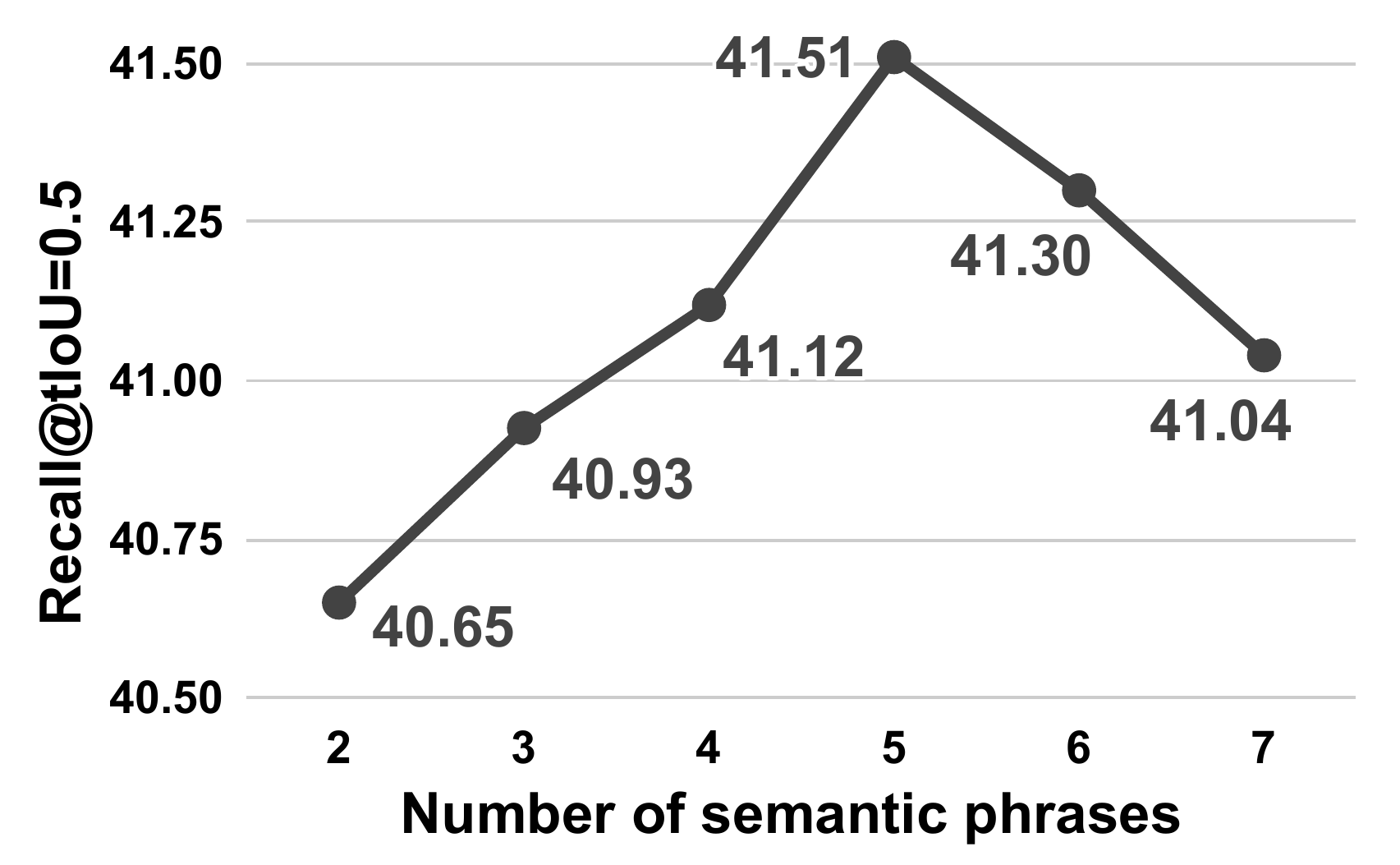}
		}
	\end{minipage}
	\vspace{0.1cm}
	\caption{
		Ablation studies with respect to the number of extracted semantic phrases.
	}
	\label{fig:natt_ablation}
\end{figure}
\begin{figure}[!t]
	\centering
	\begin{minipage}[t]{0.48\linewidth}
		\centering
		\subfloat[Charades-STA \label{tab:ablation:lambda_charades}]{
			\centering
			\includegraphics[width=\linewidth]{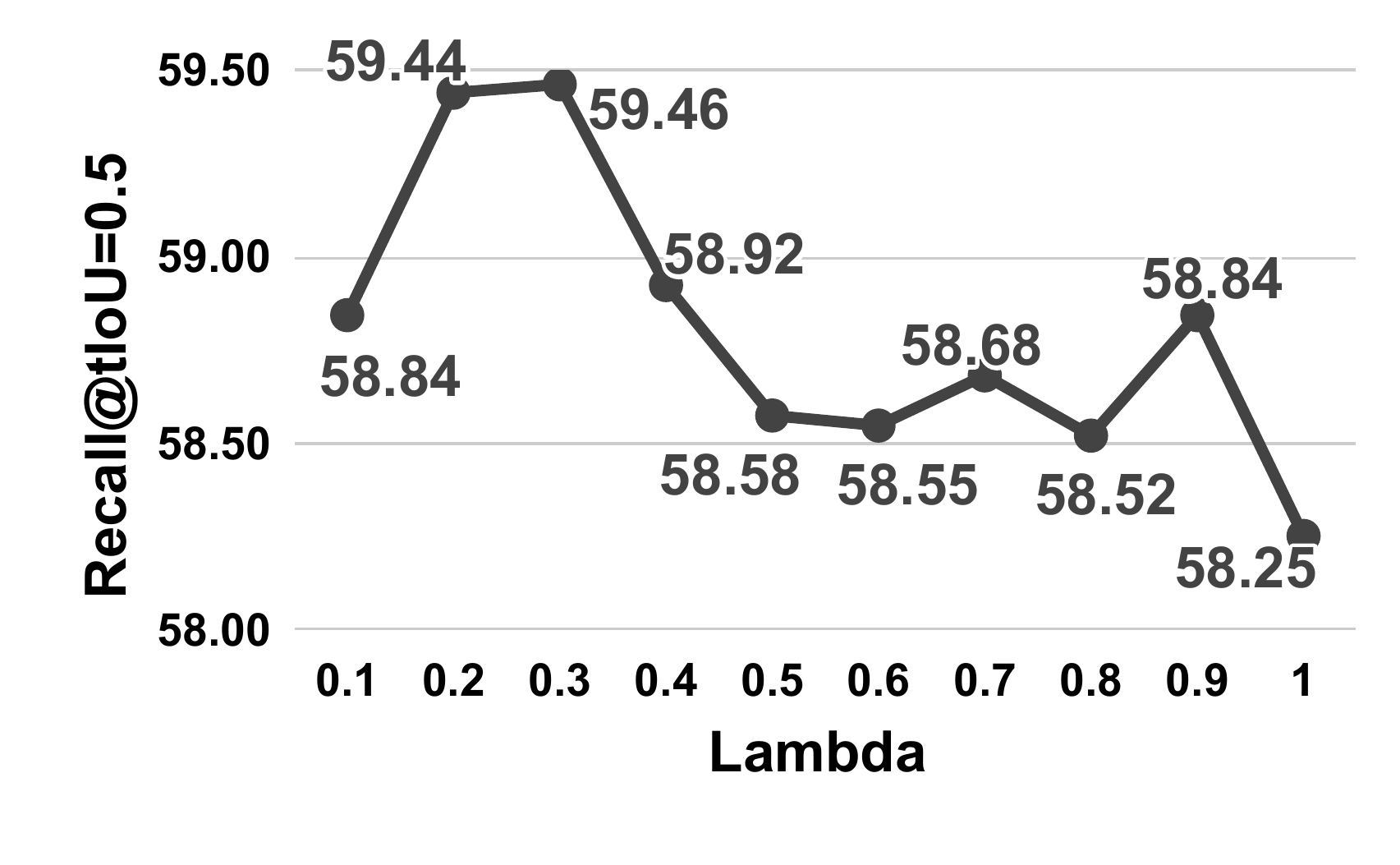}
		}
	\end{minipage}
	\begin{minipage}[t]{0.48\linewidth}
		\centering		
		\subfloat[ActivityNet Captions \label{tab:ablation:lambda_anet}]{
			\centering
			\includegraphics[width=\linewidth]{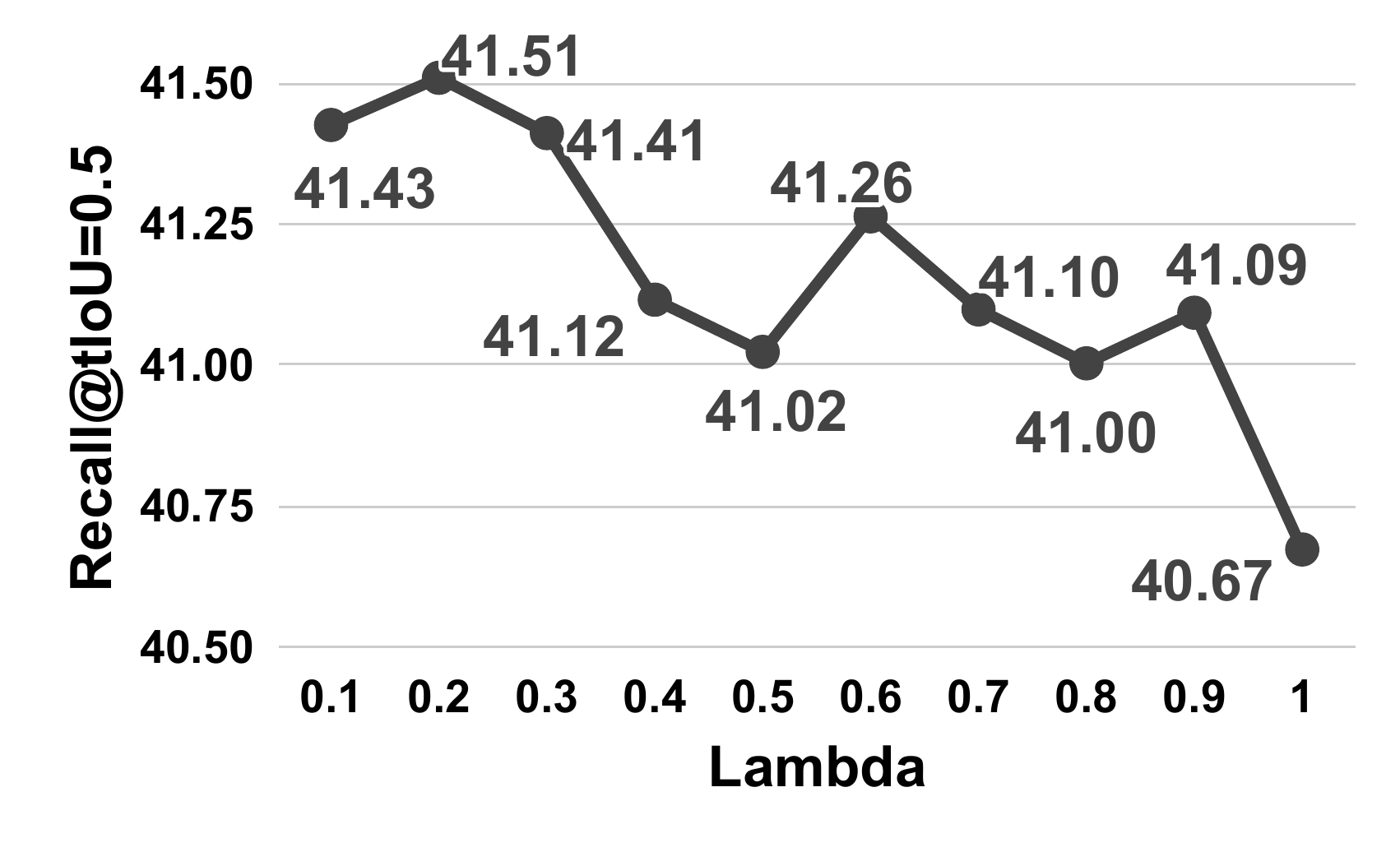}
		}
	\end{minipage}
	\vspace{0.1cm}
	\caption{
		Ablation studies across with respect to $\lambda$ values.
	}
	\label{fig:lambda_ablation}
\end{figure}

Table~\ref{tab:main_ablation} summarizes the results where we observe the following.
First, extracting semantic phrase features from the query (LGI) is more effective for precise localization than simply relying on the sentence-level representation (LGI--SQAN).
Second, regularizing the query attention weights for distinctiveness, \ie, using $\mathcal{L}_{\text{dqa}}$, enhances performance by capturing distinct constituent semantic phrases.
Third, temporal attention guidance loss $\mathcal{L}_{\text{tag}}$ improves the accuracy of localization by making models focus on segment features within the target time interval.
Finally, it is noticeable that LGI--SQAN already outperforms the state-of-the-art method at R@0.5 (\ie, 52.02\% vs.57.34\%), which shows the superiority of the proposed local-global video-text interactions in modeling relationship between video segments and a query.

\begin{table}[!t]
	\centering
	\caption{
		Performance comparison by varying the combinations of modules in local and global context modeling on the Charades-STA dataset.
		The bold-faced numbers mean the best performance.
	}
	\vspace{0.1cm}
	\scalebox{0.935}{
		\begin{tabular}{cc|c}
			\toprule
			Local Context & Global Context & R@0.5 \\
			\hline\hline
			-     &    -    & 40.86 \\
			\hdashline
			Masked NL (b=1, w=15) &    -    & 42.66 \\
			Masked NL (b=4, w=15) &    -    & 45.78 \\
			Masked NL (b=4, w=31) &    -    & 47.80 \\
			ResBlock (k=3)~~  & - & 43.95 \\
			ResBlock (k=7)~~  & - & 46.24 \\
			ResBlock (k=11)   & - & 49.78 \\
			ResBlock (k=15)   & - & 50.54 \\
			\hdashline
			
			- & NLBlock (b=1) & 48.12 \\
			- & NLBlock (b=2) & 48.95 \\
			- & NLBlock (b=4) & 48.52 \\
			\hdashline
			
			Masked NL (b=1, w=15) & NLBlock (b=1) & 50.11 \\
			Masked NL (b=4, w=15) & NLBlock (b=1) & 53.92 \\
			Masked NL (b=4, w=31) & NLBlock (b=1) & 54.81 \\
			ResBlock (k=7)~~  & NLBlock (b=1) & 55.00 \\
			ResBlock (k=15) & NLBlock (b=1) & \textbf{57.34} \\
			
			\bottomrule
		\end{tabular}
	}
	\label{tab:VTI_ablation}
\end{table}

\begin{table*}[!t]
	\centering
	\begin{minipage}[t]{0.33\linewidth}
		\centering		
		\subfloat[Performance comparison depending on the location of segment-level modality fusion in the video-text interaction. \label{tab:ablation:order}]{
			\tablestyle{11pt}{1.1}
			\centering
			\begin{tabular}{l|c}
				\toprule
				Option & R@0.5 \\
				\hline \hline
				Local-Global-Fusion & 46.96 \\
				Local-Fusion-Global & 53.47 \\
				Fusion-Local-Global & 57.34 \\
				\bottomrule
			\end{tabular}
		}
	\end{minipage}
	\begin{minipage}[t]{0.33\linewidth}
		\centering
		\subfloat[Performance comparison with respect to fusion methods. \label{tab:ablation:att_cal}]{
			\tablestyle{11pt}{1.1}
			\begin{tabular}{l|c}
				\toprule
				Option & R@0.5 \\
				\hline \hline
				Addition  & 46.75 \\
				Concatenation & 48.15 \\
				Hadamard Product & 57.34 \\
				\bottomrule
			\end{tabular}
		}
	\end{minipage}
	\begin{minipage}[t]{0.33\linewidth}
		\centering
		\subfloat[Impact of position embedding for video encoding. \label{tab:ablation:position}]{
			\tablestyle{11pt}{1.1}
			\begin{tabular}{l|c}
				\toprule
				Option & R@0.5 \\
				\hline \hline
				None & 45.70 \\
				Position Embedding & 57.34 \\
				\bottomrule
			\end{tabular}
		}
	\end{minipage}
	\caption{
		Ablations on the Charades-STA dataset.
	}
	\label{tab:additional_ablations}
\end{table*}

We also analyze the impact of two hyper-parameters in SQAN---the number of semantic phrases ($N$) and controlling value ($\lambda$) in $\mathcal{L}_{\text{dqa}}$---on the two datasets.
Fig.~\ref{fig:natt_ablation} presents the results across the number of semantic phrases in SQAN, where performances increase until certain numbers (3 and 5 for Charades-STA and ActivityNet Captions datasets, respectively) and decrease afterward. 
This is because larger N makes models capture shorter phrases and fail to describe proper semantics.
As shown in Fig.~\ref{fig:lambda_ablation}, the controlling value $\lambda$ of 0.2 and 0.3 generally provides good performances while higher $\lambda$ provides worse performance by making models focus on one or two words as phrases.

\vspace{-0.2cm}
\subsubsection{Analysis on Local-Global Video-Text Interaction}
We perform in-depth analysis for local-global interaction on the Charades-STA dataset.
For this experiment, we employ LGI--SQAN (instead of LGI) as our base algorithm to save training time.

\paragraph{Impact of local and global context modeling}
We study the impact of local and global context modeling by varying the kernel size ($k$) in the residual block (ResBlock) and the number of blocks ($b$) in Non-Local block (NLBlock).
For local context modeling, we also adopt an additional module referred to as a masked Non-Local block (Masked NL) in addition to ResBlock;
the mask restricts attention region to a local scope with a fixed window size $w$ centered at individual segments in the NLBlock.

Table~\ref{tab:VTI_ablation} summarizes the results, which imply the following.
First, the performance of the model using only segment-level modality fusion without context modeling is far from the state-of-the-art performance.
Second, incorporating local or global context modeling improves performance by enhancing the alignment of semantic phrases with the video.
Third, a larger scope of local view in local context modeling further improves performance, where ResBlock is more effective than Masked NL according to our observation.
Finally, incorporating both local and global context modeling results in the best performance gain of 16.48\% points.
Note that while the global context modeling has a capability of local context modeling by itself, it turns out to be difficult to model local context by increasing the number of NLBlocks; a combination of Masked NL and NLBlock outperforms the stacked NLBlocks, showing the importance of explicit local context modeling.

\paragraph{When to perform segment-level modality fusion}
Table~\ref{tab:additional_ablations}(a) presents the results from three different options for the modality fusion phase.
This result implies that early fusion is more beneficial for semantics-aware joint video-text understanding and leads to the better accuracy.

\paragraph{Modality fusion method}
We compare different fusion operations---addition, concatenation, and Hadamard product.
For concatenation, we match the output feature dimension with that of the other methods by employing an additional embedding layer.
Table~\ref{tab:additional_ablations}(b) shows that Hadamard product achieves the best performance while the other two methods perform much worse.
We conjecture that this is partly because Hadamard product acts as a gating operation rather than combines two modalities, and thus helps the model distinguish segments relevant to semantic phrases from irrelevant ones.

\paragraph{Impact of position embedding}
Table~\ref{tab:additional_ablations}(c) presents the effectiveness of the position embedding in identifying semantic entities at diverse temporal locations and improving the accuracy of temporal grounding.

\begin{figure}[!t]
	\centering
	\scalebox{1}{
		\includegraphics[width=\linewidth]{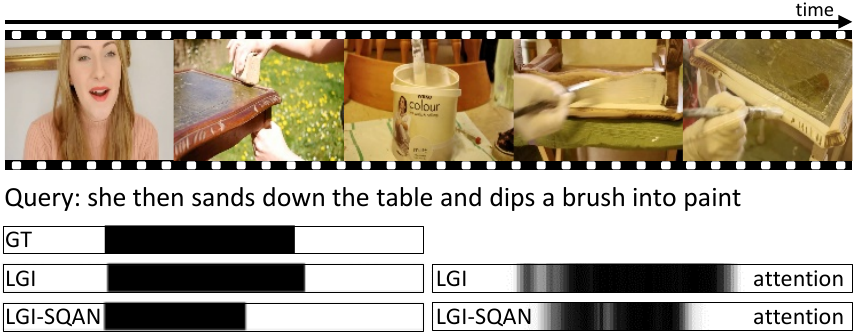}
	}
	\caption{
		Visualization of predictions of two models (LGI and LGI--SQAN) and their temporal attention weights $\ve{o}$ computed before regression.
	}
	\label{fig:pred_comp}
	\vspace{-0.2cm}
\end{figure}

\vspace{-0.2cm}
\subsubsection{Qualitative Results}
\label{sec:qualitative}
Fig.~\ref{fig:pred_comp} illustrates the predictions and the temporal attention weights $\ve{o}$ for LGI and LGI--SQAN.
Our full model (LGI) provides more accurate locations than LGI--SQAN through query understanding in a semantic phrase level, which makes video-text interaction more effective.
More examples with visualization of temporal attention weights, query attention weights $\ve{a}$ and predictions are presented in our supplementary material.

\section{Conclusion}
\label{sec:conclusion}

We have presented a novel local-global video-text interaction algorithm for text-to-video temporal grounding via constituent semantic phrase extraction.
The proposed multi-level interaction scheme is effective in capturing relationships of semantic phrases and video segments by modeling local and global contexts.
Our algorithm achieves the state-of-the-art performance in both Charades-STA and ActivityNet Captions datasets.

\vspace{-0.2cm}
\paragraph{Acknowledgments} 
This work was partly supported by IITP grant funded by the Korea government (MSIT) (2016-0-00563, 2017-0-01780), and Basic Science Research Program (NRF-2017R1E1A1A01077999) through the NRF funded by the Ministry of Science, ICT.
We also thank Tackgeun You and Minsoo Kang for valuable discussion.

{
	\small
	\bibliographystyle{ieee_fullname}
	\bibliography{egbib}
}

\section{Supplementary Material}
This supplementary document first presents the architecture of our model without semantic phrase extraction (\ie, LGI--SQAN) used for in-depth analysis on the local-global video-text interactions.
We also present additional qualitative examples of our algorithm.

\subsection{Architectural Details of LGI--SQAN}
\begin{figure}[!t]
	\centering
	\includegraphics[width=0.99\linewidth]{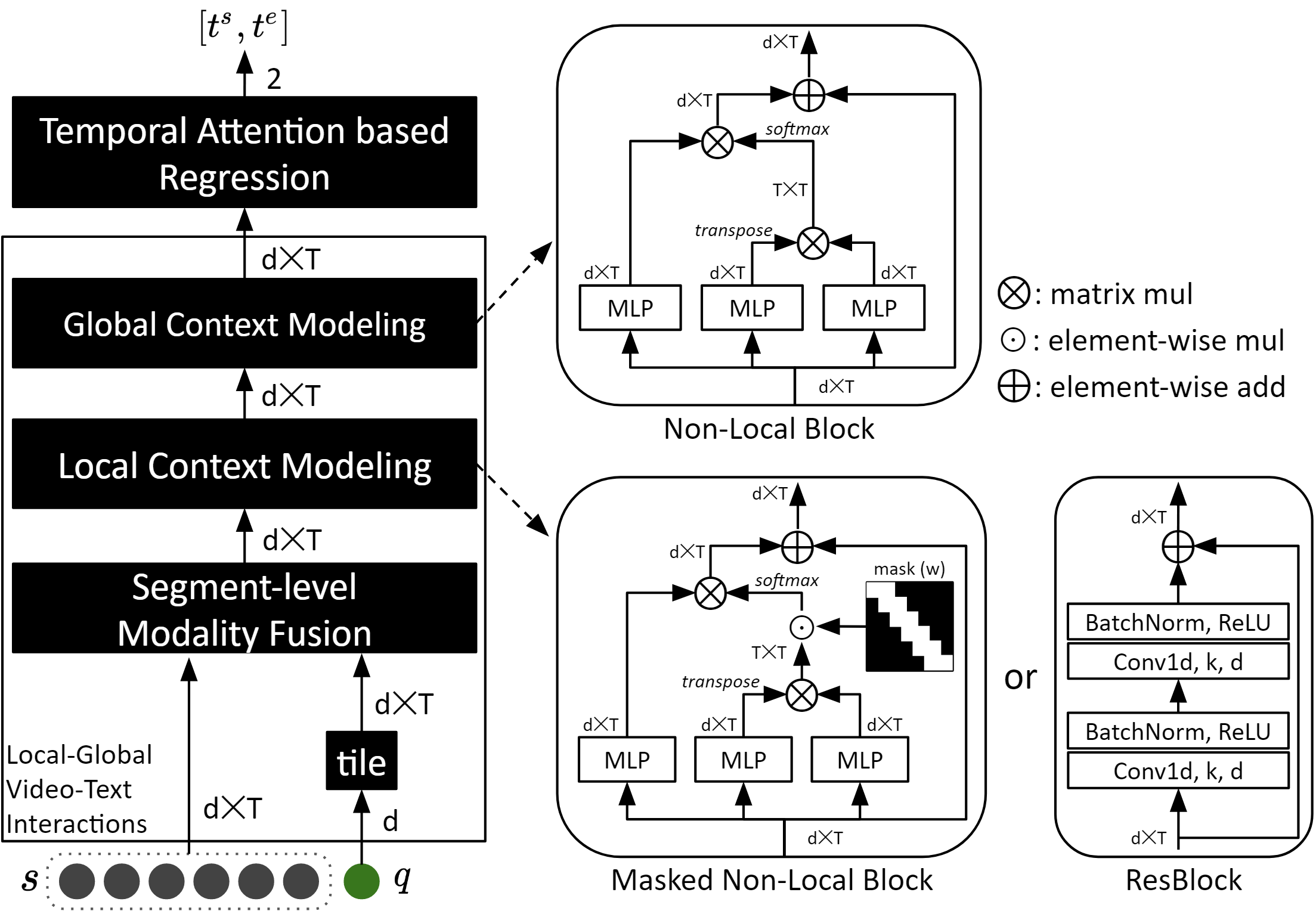}
	\vspace{0.2cm}
	\caption{
		Illustration of architecture of LGI--SQAN.
		In LGI--SQAN, we use sentence-level feature $\ve{q}$ to interact with video.
	}
	\label{fig:arch_LGI--SQAN}
\end{figure}

Compared to our full model (LGI), LGI--SQAN does not explicitly extract semantic phrases from a query as presented in Fig.~\ref{fig:arch_LGI--SQAN};
it performs local-global video-text interactions based on the sentence-level feature representing whole semantics of the query.

In our model, the sentence-level feature ($\ve{q}$) is copied to match its dimension with the temporal dimension ($T$) of segment-level features ($\ma{S}$).
Then, as done in our full model, we perform local-global video-text interactions---1) segment-level modality fusion, 2) local context modeling, and 3) global context modeling---followed by the temporal attention based regression to predict the time interval $[t^s,t^e]$.
Note that we adopt a masked non-local block or a residual block for local context modeling, and a non-local block for global context modeling, respectively.

\subsection{Visualization of More Examples}
\begin{figure*}[!t]
	\centering
	\includegraphics[width=0.99\textwidth]{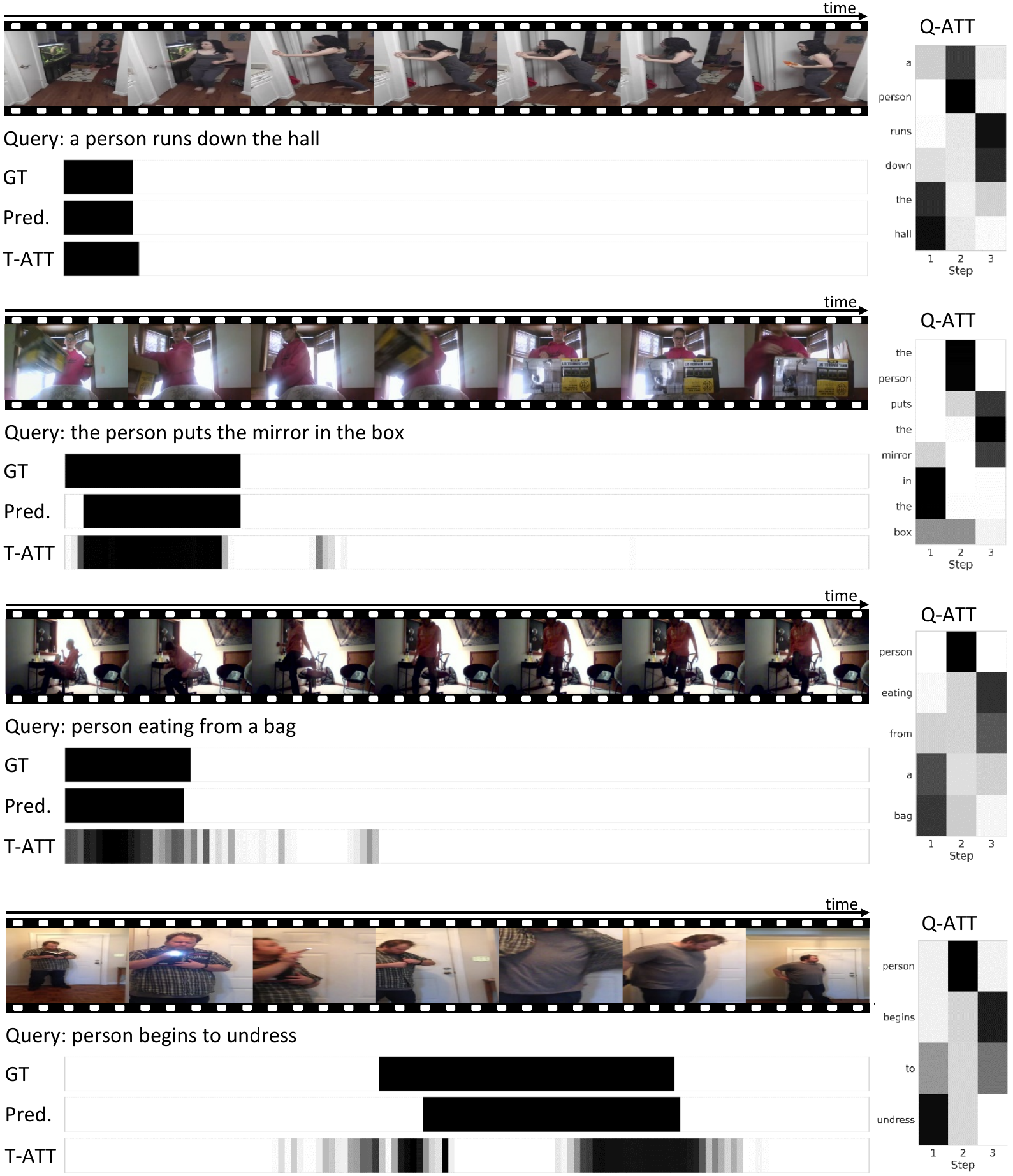}
	\vspace{0.3cm}
	\caption{
		Qualitative results of our algorithm on the Charades-STA dataset.
		T-ATT and Q-ATT stand for temporal attention weights and query attention weights, respectively.
	}
	\label{fig:success_cases_charades}
\end{figure*}

\begin{figure*}[!t]
	\centering
	\includegraphics[width=0.99\textwidth]{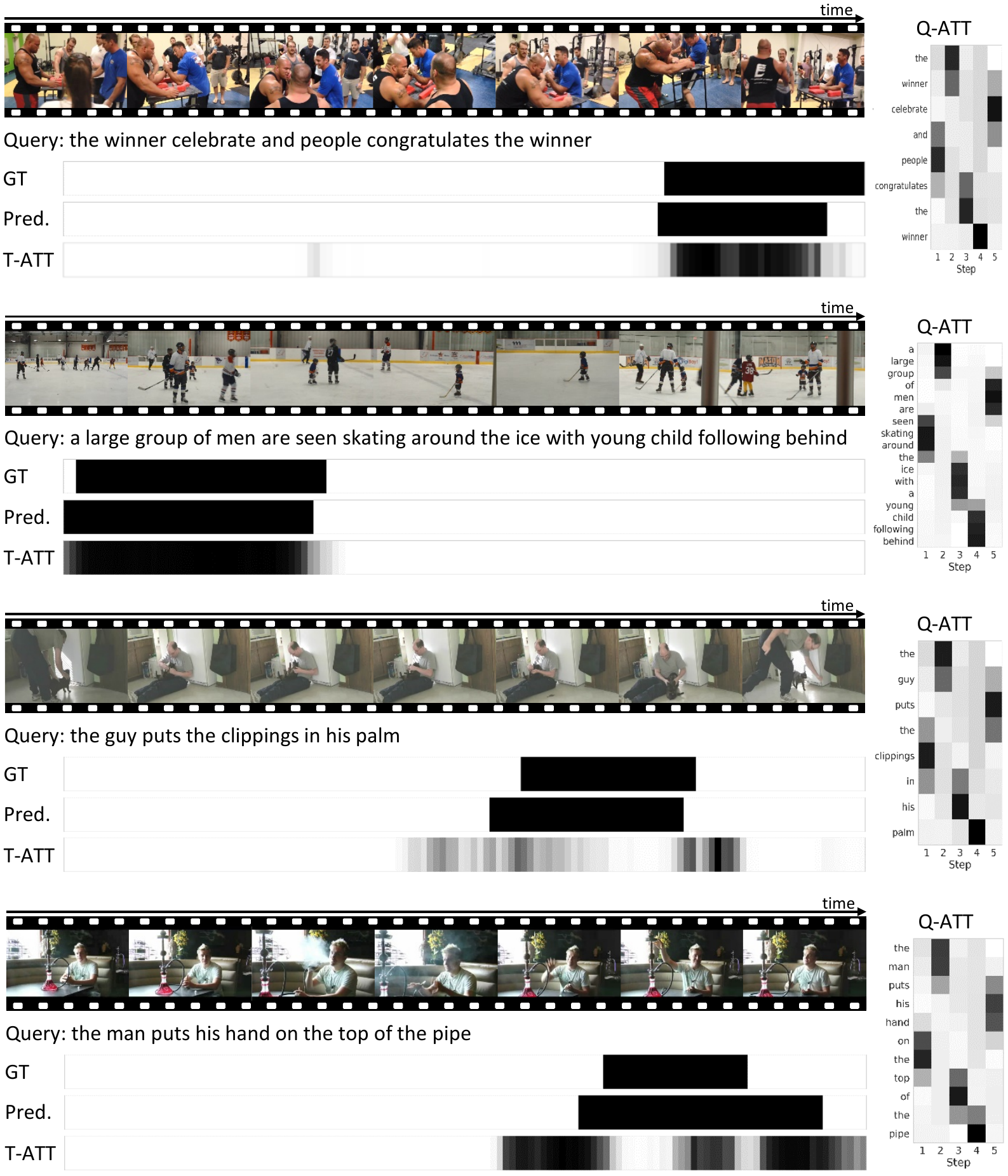}
	\vspace{0.3cm}
	\caption{
		Qualitative results of our algorithm on the ActivityNet Captions dataset.
		T-ATT and Q-ATT stand for temporal attention weights and query attention weights, respectively.
	}
	\label{fig:success_cases_anet}
\end{figure*}

Fig.~\ref{fig:success_cases_charades} and Fig.~\ref{fig:success_cases_anet} illustrate additional qualitative results in the Charades-STA and ActivityNet Captions datasets, respectively;
we present two types of attention weights---temporal attention weights $\ve{o}$ (T-ATT) and query attention weights $\ve{a}$ (Q-ATT)---and predictions (Pred.).
T-ATT shows that our algorithm successfully attends to relevant segments to the input query while Q-ATT depicts that our sequential query attention network favorably identifies semantic phrases from the query describing actors, objects, actions, etc.
Note that our model often predicts accurate time intervals even from the noisy temporal attention.

\begin{figure*}[h]
	\centering
	\includegraphics[width=0.99\linewidth]{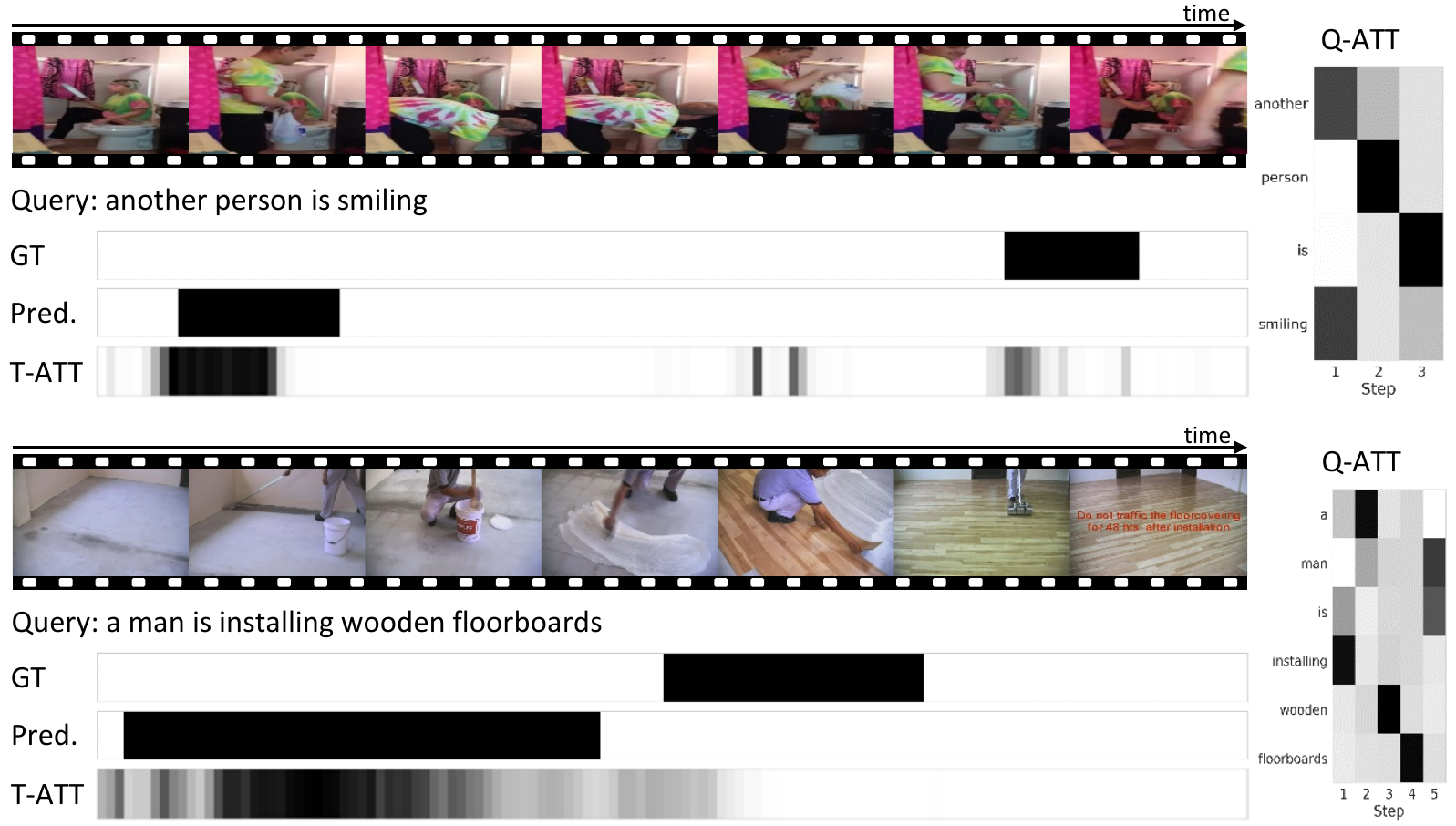}
	\vspace{0.3cm}
	\caption{
		Failure case of our algorithm.
		Examples in the first and second row are obtained from the Charades-STA and Activity Captions datasets, respectively.
	}
	\label{fig:failure_cases}
\end{figure*}

Fig.~\ref{fig:failure_cases} demonstrates the failure cases of our algorithm.
As presented in the first example of Fig.~\ref{fig:failure_cases}, our method fails to localize the query on the confusing video, where a man looks like smiling at multiple time intervals.
However, note that the temporal attention of our method captures the segments relevant to the query at diverse temporal locations in a video.
In addition, as presented in the second example of Fig.~\ref{fig:failure_cases},
our model sometimes fails to extract proper semantic phrases;
`wooden' and `floorboards' are captured at different steps although `wooden floorboards' is more natural, which results in the inaccurate localization.

~\\ \\ \\ \\ \\ \\ \\ \\ \\ \\ \\ \\

~\\ \\ \\ \\ \\ \\ \\ \\ \\ \\ \\ \\
~\\ \\ \\ \\ \\ \\ \\ \\ \\ \\ \\ \\
~\\ \\ \\ \\ \\ \\ \\ \\ \\ \\ \\ \\

\end{document}